\documentclass[letterpaper, 10 pt, conference]{ieeeconf}  % Comment this line out if you need a4paper

\IEEEoverridecommandlockouts                              % This command is only needed if 
\usepackage{color}
\usepackage{amsmath, amssymb, amscd, amsfonts} % assumes amsmath package installed
\usepackage{graphics} % for pdf, bitmapped graphics files
\usepackage{graphicx}
\usepackage{hyperref}
\usepackage{tabularx}
\usepackage{multirow}
\usepackage{vcell}
\usepackage{colortbl}
\usepackage{booktabs}
\usepackage{cite}
\usepackage{float}
\usepackage{subfig}
\usepackage{caption}
\usepackage{algorithm}
\usepackage{algorithmic}
\usepackage{bbding}
\usepackage{array}
\usepackage{makecell}
\usepackage{lipsum}
\usepackage{tikz}
\usepackage{url}
\usetikzlibrary{shapes.geometric, arrows, positioning}

\UseRawInputEncoding

\newcommand{\method}{\textit{ControlMTR}}

\graphicspath{ {./figs/} }

\title{\LARGE \bf
ControlMTR: Control-Guided Motion Transformer with Scene-Compliant Intention Points for Feasible Motion Prediction
}

\author{Jiawei Sun$^{1}$, Chengran Yuan$^{1}$, Shuo Sun$^{1}$, Shanze Wang$^{1}$, Yuhang Han$^{1}$,\\ Shuailei Ma$^{2}$, Zefan Huang$^{1}$, Anthony Wong$^{3}$, Keng Peng Tee$^{3}$,  Marcelo H. Ang Jr.$^{1}$% <-this % stops a space
\thanks{
    $^{1}$Jiawei Sun, Chenran Yuan, Shuo Sun, Shanze Wang, Yuhang Han, Zefan Huang and Marcelo H. Ang Jr. are with the Department of Mechanical Engineering, National University of Singapore, Singapore 119077 (e-mail: \{sunjiawei, chengran.yuan, shuo.sun, shanze.wang, yuhang\_han, huangzefan \}@u.nus.edu; mpeangh@nus.edu.sg).
    $^{2}$Shuailei Ma is with College of Information Science and Engineering, Northeastern University, Shenyang, China, 110819 (e-mail: xiaomabufei@gmail.com).
    $^{3}$Keng Peng Tee, Anthony Wong are with Moovita Pte Ltd, Singapore, 599489 (e-mail:\{ anthonywong, kptee\}@moovita.com).
}
    }
\begin{document}

\maketitle
\thispagestyle{empty}
\pagestyle{empty}

%%%%%%%%%%%%%%%%%%%%%%%%%%%%%%%%%%%%%%%%%%%%%%%%%%%%%%%%%%%%%%%%%%%%%%%%%%%%%%%%
\begin{abstract}
The ability to accurately predict feasible multi-modal future trajectories of surrounding traffic participants is crucial for behavior planning in autonomous vehicles. 
The Motion Transformer (MTR) \cite{shi_mtr_2022}, a state-of-the-art motion prediction method, alleviated mode collapse and instability during training and enhanced overall prediction performance by replacing conventional dense future endpoints with a small set of prior motion intention points. 
However, the fixed prior intention points make the MTR multi-modal prediction distribution over-scattered and infeasible in many scenarios.
In this paper, we propose the {\method} framework to tackle the aforementioned issues by generating scene-compliant intention points and additionally predicting driving control commands, which are then converted into trajectories by a physics-based model with soft kinematic constraints. 
These control-generated trajectories will guide the directly predicted trajectories by an auxiliary loss function. Together with our proposed scene-compliant intention points, they can effectively restrict the prediction distribution within the road boundaries and suppress infeasible off-road predictions while enhancing prediction performance. 
Remarkably, without resorting to additional model ensemble techniques, our method surpasses the baseline MTR model across all performance metrics, achieving notable improvements of 5.22\% in SoftmAP and a 4.15\% reduction in MissRate. Significantly, our method also achieves a 41.5\% reduction in the MTR's cross-boundary rate, ensuring that the prediction distribution remains within the drivable area.
% In addition, our method explicitly models the relative historical movements between agents and map polylines as a novel input modality and applies the Multi-Scale Noding (MSG) and Multi Context Gating (MCG) module to learn their intricate spatial and temporal interactions. 
% Multi-Context Gating (MCG) module and local attention transformer module are utilized for efficiently learning both global and local structures for the traffic-scene graph. 

\end{abstract}

%%%%%%%%%%%%%%%%%%%%%%%%%%%%%%%%%%%%%%%%%%%%%%%%%%%%%%%%%%%%%%%%%%%%%%%%%%%%%%%%
\section{INTRODUCTION}

Accurately and efficiently predicting surrounding traffic participants’ future possible trajectories plays a pivotal role in autonomous driving systems. It helps the self-driving agents better understand the current scenarios, thus making safe and reliable decisions and giving comfortable planning commands. The motion prediction problem is still unsolved, in part due to the highly uncertain and stochastic nature of other road users' future motion, especially in the long-term prediction field, as well as the complex interrelated multi-modal context features comprising road polylines, agents' historical states, and traffic light states \cite{varadarajan_multipath_2021}.

\begin{figure}[t]
\centering
\includegraphics[width=\linewidth]{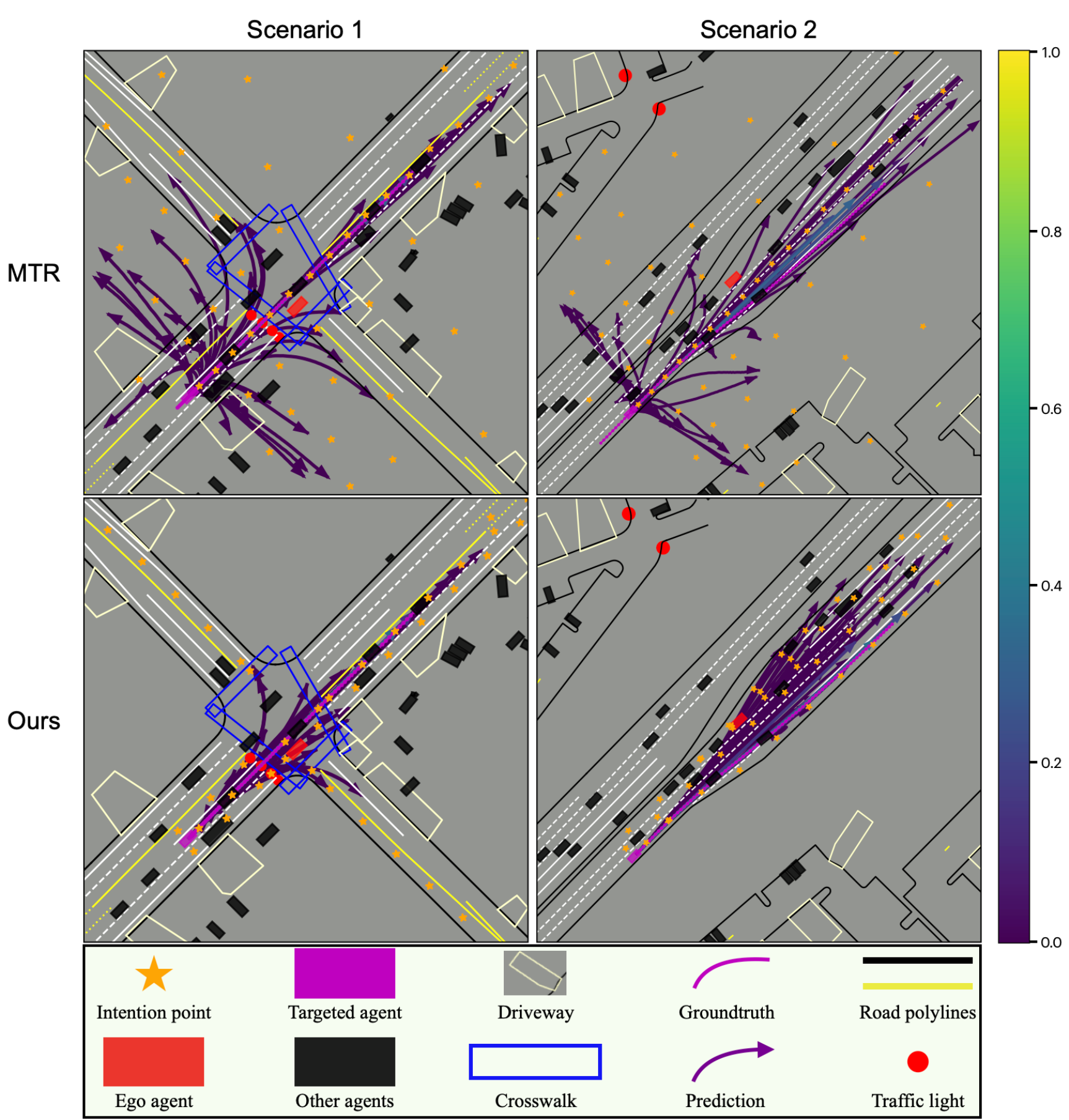}
\caption{
    A visual comparison between the motion prediction results of MTR and our framework. The predicted top-64 trajectories for each agent are colored from yellow (high) to purple (low) according to their associated probability. Our prior intention points and their associated trajectories are generated in a more compliant fashion that adheres to different lane geometries and traffic rules instead of being restricted by a fixed pattern (MTR).
}
\label{fig:graphical-abstract}
\end{figure}

In order to produce multi-modal prediction outputs, previous works can be generally divided into two different decoder pipelines: one-stage direct-regression prediction \cite{huang2023differentiable} and two-stage endpoints-conditioned prediction\cite{gu_densetnt_2021}. The previous direct-regression methods suffer from mode collapse and training instability issues, mainly due to the imbalanced dataset scenario distribution. The other type of pipeline first tries to predict a set of possible endpoints and then regresses the full trajectories conditioned on the predicted endpoints. This strategy improves the trajectories' diversity by predicting all potential candidate endpoints. However, such methods' accuracy demands a large number of potential endpoints, which leads to a heavy computation cost. 

Recently, MTR \cite{shi_mtr_2022} proposes a novel decoder structure that adopts a small set of motion query pairs to jointly optimize the global possible intentions and local agent movements. In particular, MTR utilizes 64 fixed prior intention points for each target agent and predicts 64 corresponding trajectories. The intention points are generated by K-Means algorithms on all agents' ground truth endpoints, and each intention point is viewed as a potential motion mode. This approach reduces prediction uncertainty, but the fixed nature of these intention points may not align well with the possible trajectory distributions when the scenario changes. As shown in Fig. \ref{fig:graphical-abstract}, most intention points fall on the opposite lane or even outside the road boundaries, indicating a limitation in their adaptability to varied scenarios. Such limitation makes MTR fail to output enough feasible trajectories, and most of its predictions disobey the traffic rules. Furthermore, MTR treats all agents' tracks and the map as polylines and employs a Pointnet-like structure to aggregate their features. However, this straightforward approach might not be effective in capturing the complex interactions between agents and map polylines.

Given the drawbacks of MTR, we propose our {\method} framework. Our model explicitly considers historical relative movements between agents and road boundary constraints as an additional modality input. We propose a Multi-Scale GRU(MSG) module to better encode the temporal information and adopt a cascading Multi-Context Gating(MCG) module for better inter-class feature fusion. To confine the predicted trajectories within road boundaries, we add control guidance for direct prediction results and adopt scene-compliant intention points as the static intention query. The experiments show that our method outperforms MTR in all prediction metrics while being able to produce abundant feasible on-road future trajectories.

    The contributions of this paper can be summarized as follows:
    \begin{itemize}
        \item We propose a method that can generate traffic-rule-compliant intention points as the static intention query and output additional control commands to restrict the direct prediction results, which ensures the predicted trajectories adapt to real-world map topologies, effectively mitigating off-road and infeasible predictions. 
        \item We design a novel scene encoder architecture that explicitly considers historical relative movements between agents and map constraints as an extra modality input. MSG module is adopted for enhanced temporal encoding, and MCG module is applied for the efficient probing of global interactions among scene elements.
        \item  
        Evaluation results demonstrate that our method outperforms MTR in prediction benchmarks and dramatically reduces the rate of off-road predictions from 66.13\% to 38.68\%. It achieves comparable performance with the top-performing methods on the Waymo  Motion Prediction Leaderboard. 
    \end{itemize}

\section{Related Work}

\subsection{Traffic Scene Representation}

Addressing motion prediction necessitates robust representations of traffic scene components, particularly high-definition map elements and agents' historical trajectories. 
Many prior works \cite{ chai_multipath_2019, gilles_home_2021,rEcOAT} rasterized the entire scene into a Bird's-Eye-View (BEV) image-like tensor as the model input and employed Convolutional Neural Networks (CNN) \cite{tan_efficientnet_2020, konev2022motioncnn} to encode and extract key information. 
Each channel of the BEV image represents a different type of map element, such as lane markers, road boundaries, traffic signals, etc. Meanwhile, agent trajectories are either added as additional image channels \cite{chai_multipath_2019, cui_multimodal_2019} or separately processed by temporal neural networks such as the Recurrent Neural Networks (RNN) \cite{alahi_social_2016}. 
However, despite the fact that rasterized methods provide unified representations for different elements in a dense, fixed-sized data structure, they are limited by their resolution and contain information loss caused by quantization errors.
Recently, vectorized methods \cite{gao_vectornet_2020, liang_learning_2020, huang2023differentiable} have gained traction for their sparse encoding efficiency and prowess in capturing intricate structural information. Unlike rasterized methods, these methods vectorize the scene as a set of entities with semantic and geometric attributes, emphasizing their interrelations. 
VectorNet \cite{gao_vectornet_2020} pioneered the direct incorporation of vectorized information of both lanes and agents, which has influenced subsequent research \cite{ngiam_scene_2022, sun2022m2i, zhou_query-centric_nodate}. The vectorized input approach can be further categorized into three different normalization methods: agent-centric \cite{ngiam_scene_2022}, scene-centric \cite{sun2022m2i}, and query-centric \cite{zhou_query-centric_nodate}.

In our model, we utilize a vectorized input representation and apply an agent-centric normalization method. Historical relative movements between agents and map polylines are explicitly considered as additional modality input. We further utilize the MSG module to aggregate the input temporal information and Pointnet to encode spatial information.

\begin{figure*}[ht]
    \centering
    \includegraphics[width=13
    cm]{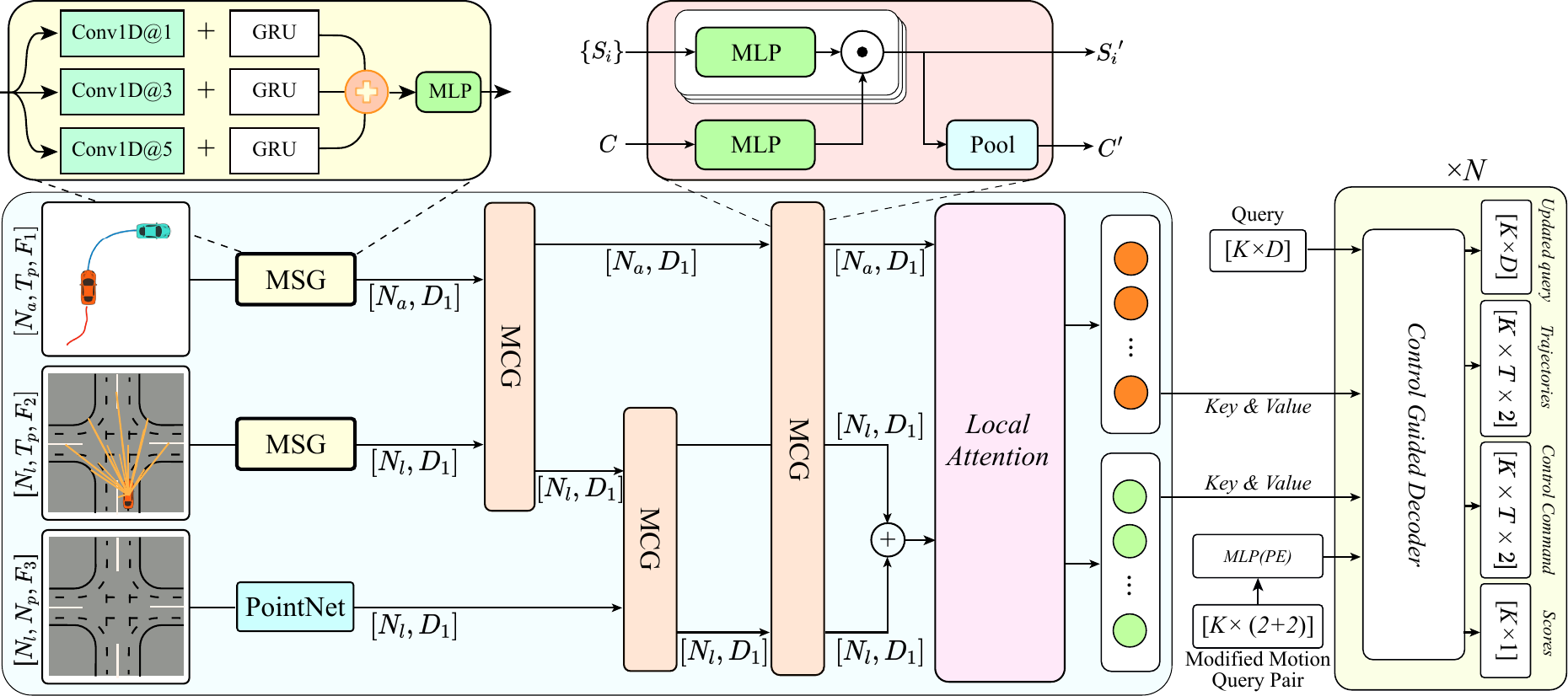} 
    \caption{Encoder structure of our proposed method.}
    \label{fig:encoder-structure}
    \vspace{-1em}
\end{figure*}

\subsection{Multi-modality Modeling}

Given the encoded scene context features, previous research has delved into various approaches for forecasting multi-modal future motions. 
mmTransformer \cite{liu_multimodal_2021} proposed a region-based training method, which ensures that each proposal embodies a distinct pattern. 
In contrast, goal-based forecasting methods \cite{fang_tpnet_2021,gu_densetnt_2021} predict multiple agent endpoints, then generate a complete trajectory for each selected goal. 
Popular methods such as DenseTNT \cite{gu_densetnt_2021} utilized a trajectory prediction model to derive a goal probability distribution, which is subsequently transformed into a set of goals by a predictor. 
Meanwhile, some research focuses on predicting trajectories using encoded agent attributes \cite{ngiam_scene_2022}, heatmaps \cite{gilles_home_2021, gilles_gohome_2021}, or latent anchor embeddings \cite{varadarajan_multipath_2021}. 
However, goal-based strategies often suffer from efficiency challenges due to numerous goal candidates, while direct-regression methods tend to converge at a slower pace, given that many motion predictions can originate from the same agent features.
Recently, MTR \cite{shi_mtr_2022} mitigates the aforementioned issues by using a small set of fixed prior intention points to help reduce the prediction uncertainty while maintaining high prediction diversity. Its successor， MTR++ \cite{shi_mtr++_2023}， leverages query-centric input representation and mutually-guided intention queries to efficiently model multi-agent interactions. And LLM-augmented MTR\cite{zheng2024large} utilizes a large language model to enhance the model's reasoning ability. Other methods like MacFormer \cite{feng_macformer_2023} consider map constraints by applying a coupling module to tightly combine agent and map features, optimizing predictions through a reference extractor and multi-task strategy. 

In our framework, scene-compliant prior intention points are utilized in the decoder, and additional control commands are generated with soft kinematic constraints to guide the direct prediction trajectories by adding an auxiliary loss function. 
Together, these two innovations ensure the model is able to generate abundant feasible predictions.

% \subsection{Transformer}

\section{Methodology}

\subsection{Problem Formulation}
In a particular traffic scenario, there are $N_{a}$ interactive agents represented as $\mathcal{A} = \{{A}_{1}, {A}_{2}, \ldots, {A}_{N_{a}}\}$, along with contextual map information denoted as $\mathcal{M}$. The historical trajectory of agent $A_{i}$ for the last $T_{p}$ time intervals is recorded as $A_{i}=\{a_{i}^{t-T_{p}},\ldots, a_{i}^{t}\}$, where $a^t_i \in \mathbb{R}^{F_1}$ are the agent attributes at each time step, such as position, orientation and velocity. The map data $\mathcal{M}$ encompasses various types of road features, such as centerlines, road boundaries, roadside edges, crosswalks, and speed bumps, among others. All these road representations are uniformly subdivided as $\mathcal{M}=\{{m}_{1}, {m}_{2}, \ldots, {m}_{N_l}\} $ and contain an equal number of waypoints $m_{i}=\{{p}_{1}, {p}_{2}, \ldots, {p}_{N_p}\}$, where $p_i \in \mathbb{R}^{F_2}$ are the waypoint attributes consisting of the current waypoint coordinates, the direction towards the next waypoint and etc. The predictor $\mathcal{P}_{\theta }$ takes in both map context $\mathcal{M}$ and agents' historical states $\mathcal{A}$ and outputs $K$ future potential trajectories for each target agent over the future $T_{f}$ timesteps with corresponding probability $p_{k}$.

\subsection{Input Representation}
An agent-centric approach is adopted to standardize all inputs relative to the current state of the target agent. 
% To simplify, we assume that there is only one target agent, allowing us to introduce our method clearly.
For each target agent, its past $T_{p}$ states of surrounding $N_{a}$ agents (including itself) can be represented as $\mathcal{A} \in \mathbb{R}^{N_{a}\times T_{p}\times F_{1}}$, while an input map tensor $\mathcal{M}\in \mathbb{R}^{N_{l}\times N_{p}\times F_{2}}$ can be assembled from $N_{l}$ number of map polylines. 
Consequently, the relative movement between the target agent and road polylines can be represented as $\mathcal{R}\in \mathbb{R}^{N_{l}\times T_{p}\times F_{3}}$, where $F_{3}$ is indicative of the relative position and orientation between the target agent and center of each road polyline over the past timesteps.
\subsection{Network Structure}
For the encoder part, we first utilize MSG and Pointnet-like modules to aggregate spatial and temporal inputs into tokenized features and then apply MCG and local attention to probe both global and local connections between each token feature. In the decoder part, we follow \cite{shi_mtr_2022}'s pipeline but use our proposed scene-compliant intention points (Algorithm \ref{Intention Points Generation}) as static intention query and generate additional driving control commands. The control commands will be converted into trajectories using a simple kinematic model (Algorithm \ref{algorithm-motion-update-vectorized}) to guide the direct prediction results further.

\textbf{Feature Extraction.}
Time series information, including agent historical states $\mathcal{A}$ and relative movement $\mathcal{R}$, is processed using a Multi-Scale GRU (MSG) model, which is a network designed to discern temporal relationships. Initially, the data is concurrently passed through a 1D CNN module with varying kernel sizes of $1$, $3$, and $5$. Subsequently, it progresses through a two-layer GRU network, whose output at the final timestep is captured, concatenated across feature dimensions, and subsequently passed through an additional MLP layer to generate the final feature token (see Fig. \ref{fig:encoder-structure}). For the road polylines data $\mathcal{M}$, a simple PointNet-like network is employed to extract the spatial feature of each polyline:

\begin{equation}
\begin{array}{clcl}
    \mathcal{A}_{1} &= \text{MSG}(\mathcal{A}), &\mathcal{A}_{1} \in \mathbb{R}^{N_{a}\times D_{1}}, \\
    \mathcal{R}_{1} &= \text{MSG}(\mathcal{R}), &\mathcal{R}_{1} \in \mathbb{R}^{N_{l}\times D_{1}}, \\
    \mathcal{M}_{1} &= \phi(\text{MLP}(\mathcal{M})), &\mathcal{M}_{1}\in \mathbb{R}^{N_{l}\times D_{1}},
\end{array}
\end{equation}
where $\phi(\cdot)$ denotes the max-pooling operation.

\begin{figure*}[ht]
    \centering
    \includegraphics[width=13
    cm]{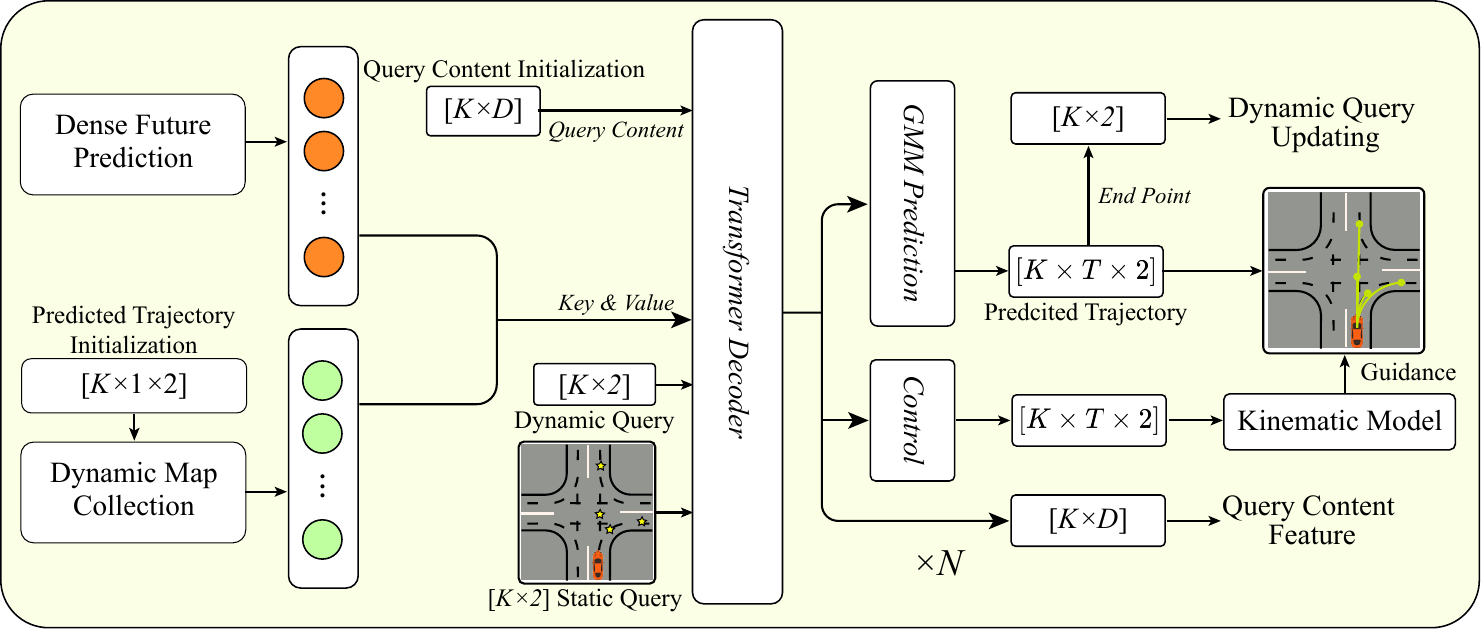} 
    \caption{Details of our decoder architecture.}
    \label{fig:decoder-structure}
    \vspace{-1em}
\end{figure*}

\textbf{Feature Fusion.}
In this process, we integrate encodings from various input modalities using Multi-Context Gating (MCG) as proposed in \cite{varadarajan_multipath_2021}. 
This mechanism works similarly to self-attention but with greater efficiency, and each input token has the capacity to interact with all other tokens. 
Meanwhile, we employ a cascading method where, at each stage, two distinct modalities are selected from a set of three to be input into the MCG module. 
The output generated by one MCG module is then fed into the next MCG module in the sequence as illustrated in Fig. \ref{fig:encoder-structure}.

\begin{equation}
\begin{split}
    (\mathcal{A}_{2}, \mathcal{R}_{2}) &= \text{MCG}(\mathcal{A}_{1}, \mathcal{R}_{1}), \\
    (\mathcal{M}_{2}, \mathcal{R}_{3}) &= \text{MCG}(\mathcal{M}_{1}, \mathcal{R}_{2}), \\
    (\mathcal{A}_{3}, \mathcal{M}_{3}) &= \text{MCG}(\mathcal{A}_{2}, \mathcal{M}_{2}).
\end{split}  
\label{eqation:mcg}
\end{equation}

Next, the relative movement feature $\mathcal{R}{3}$ and the road polylines feature $\mathcal{M}_{3}$ are added together to form the map context feature $\mathcal{M}_{map} =\mathcal{M}_{3}+\mathcal{R}_{3}$. Meanwhile, $\mathcal{A}_{3}$ is considered as the agent feature $\mathcal{A}_{agent}$.

Grasping the nuanced interactions between the target agent and adjacent map polylines is crucial for precise forecasting. Therefore, we use the K-Nearest Neighbor (KNN) algorithm to embed prior position information into our context encoder through the implementation of local attention strategies. The $i_{th}$ transformer encoder layer can be formulated as \cite{shi_mtr_2022}:

\begin{equation}
    \begin{split}
        Q^{i} = \text{MHA} (
            & Q^{i-1}+\text{PE}({Q^{i-1}}), \\ 
            & \mathcal{K}(Q^{i-1}) + \text{PE}({\mathcal{K}(Q^{i-1}))}, 
            \mathcal{K}(Q^{i-1}) 
        ),
    \end{split}
\end{equation}
where $\text{MHA}(\cdot,\cdot,\cdot)$ represents the multi-head attention function, $Q^{0}=[\mathcal{M}_{map},\mathcal{A}_{agent}]\in \mathbb{R}^{(N_{a}+N_{m})\times D_{1}}$, and
$\mathcal{K}(\cdot)$ denotes the $K$-nearest neighbours (KNN) algorithm, which is used to identify the $K$ nearest polylines relative to each query. The term $\text{PE}(\cdot)$ refers to the sinusoidal positional encoding assigned to input tokens, incorporating the most recent position of each agent and the central point of each map polyline. The final output of the local attention layer will be viewed as context encoding for both map and agents $[\mathcal{M}_{map},\mathcal{A}_{agent}]=Q^{N}$.

\textbf{Control Guided Decoder.} We adopt a similar decoder structure as MTR \cite{shi_mtr_2022} but with some key differences in terms of generating scene-compliant intention points and using control commands to constrain direct prediction results (Fig. \ref{fig:decoder-structure}).
% , and in this part, we will omit the details of the similar parts with the MTR model. 

First, the encoder's outputs, namely the agent feature $\mathcal{A}_{agent}$ and the map feature $\mathcal{M}_{map}$, are fed into the dense future prediction and the dynamic map collection modules, respectively. The former provides a prior estimate of the agents' futures, while the latter dynamically selects the features from the nearest $L$ map polylines based on the previous decoder layer's predicted trajectories: 
\begin{equation}
    \begin{split}
        \Pi_{dense} &= \text{MLP}(\mathcal{A}_{agent})\in \mathbb{R}^{N_{a}\times(T\times4)},
        \\
        \mathcal{A}_{agent} &= \text{MLP}([\mathcal{A}_{agent},\text{MLP}(\mathcal{F}_{pred})]) \in \mathbb{R}^{N_{a}\times D},
        \\
        \mathcal{M}_{map} &= \varphi(\mathcal{M}_{map}) \in \mathbb{R}^{L\times D} ,L \ll N_{m},
    \end{split}
\end{equation}
where $\Pi_{dense}$ is the dense future prediction result for agents. $\varphi(\cdot)$ denotes dynamic map collection operation.
For each agent, we predict one future motion for their position and velocity along future $T$ timesteps.

Unlike MTR, which utilized fixed intention points generated by the K-Means algorithm on all endpoints of ground truth trajectories, we propose a method for generating scene-compliant intention points $P_{static}$ that can dynamically adapt to various map geometries, as described in Algorithm \ref{Intention Points Generation}. Given the current position of the agent, a traffic-rule complaint breath-first search is utilized to find all permissible successive road centerlines based on the road graph. Subsequently, ${K}$ number of waypoints can be uniformly sampled along the selected centerlines as our intention points, which are guaranteed to be valid and traffic rule compliant. In this way, a large number of intention points that used to be off-road or invalid can now be relocated to meaningful positions, helping the network accurately pinpoint the correct global intention and further generate feasible predictions. 

\begin{algorithm}[h]
    \caption{Intention Points Generation}
    \label{Intention Points Generation}
    \begin{algorithmic}[1]
        \REQUIRE ~~\\ 
            Starting lane $l_{start}$; \\
            Set of visited lanes $\mathcal{L}_{visited}$; \\
            Frontier queue $Q_{frontier}$; \\
            Priority queue for selected waypoints $Q_{p}$; \\
            Maximum search distance $d_{max}$; \\
            Maximum number of lanes $N_{l}$; \\
            Number of intention points $N_{p}$. \\
        \ENSURE ~~\\ 
            Sampled Intention Points $\mathcal{P}$.
            \STATE $\mathcal{L}_{visited} \gets \O$
            \STATE $Q_{frontier}.push(l_{start})$ 
            \STATE $\mathcal{L}_{visited}.insert(l_{start})$ 
            \WHILE{$Q_{frontier} \neq \O$ \AND $len(\mathcal{L}_{s}) \leq N_{l}$}
                \STATE $l_{curr} \gets pop(Q_{frontier})$
                \STATE $\mathcal{L}_{ch, nb} \gets children(l_{curr}) \cup neighbors(l_{curr})$
                \FORALL {$l \in \mathcal{L}_{ch, nb}$}
                    \IF{$l \notin \mathcal{L}_{visited}$ \AND $distance(l,l_{start}) \leq d_{max}$}
                        \STATE $\mathcal{L}_{visited}.insert(l)$
                        \STATE $Q_{frontier}.push(l)$
                        \FORALL{waypoint $p \in l$}
                            \STATE $Q_{p}.push(p, distance(p, l_{start}))$
                        \ENDFOR
                    \ENDIF
                \ENDFOR
            \ENDWHILE
            \RETURN $\mathcal{P} \gets sample(Q_{p}, number = N_{p})$
    \end{algorithmic}
\end{algorithm}

For the $j_{th}$ decoder layer, we first use self-attention to propagate information between ${K}$ different motion modes.

\begin{equation}
    \begin{split}
        Q^{j} = \text{MHA} (
            & Q^{j-1} + \text{MLP}(\text{PE}({P_{static}})), \\ 
            & Q^{j-1} + \text{MLP}(\text{PE}({P_{static}})), 
            Q^{j-1}) 
        ) \in \mathbb{R}^{K\times D},
    \end{split}
\end{equation}
where $Q^{0}\in\mathbb{R}^{K\times D}$ is initialized to be all zeros.

Then, two cross-attention layers are adopted respectively for aggregating features from the output of dense future prediction $\mathcal{A}_{agent}$ and dynamic map collection $\mathcal{M}_{map}$.
\begin{equation}
    \begin{split}
        Q^{j}_{agent} = \text{MHA} (
            & Q^{j} + \text{MLP}(\text{PE}({P^{j}_{dyn}})), \\ 
            & [\mathcal{A}_{agent}, \text{PE}(\mathcal{A}_{agent})], 
            \mathcal{A}_{agent}) 
        ) \in \mathbb{R}^{K\times D},
        \\
        Q^{j}_{map} = \text{MHA} (
            & Q^{j}+\text{MLP}(\text{PE}({P^{j}_{dyn}})), \\ 
            & [\mathcal{M}_{map},\text{PE}(\mathcal{M}_{map})], 
            \mathcal{M}_{map}) 
        )\in \mathbb{R}^{K\times D},
    \end{split}
\end{equation}
where ${P^{0}_{dyn}}\in\mathbb{R}^{K\times2}$ is initialized as $P_{static}$ and ${P^{j}_{dyn}}$ is updated by the endpoint of the $(j-1)$-th decoder's predicted trajectory. 
$[\cdot,\cdot]$ denotes the concatenation operation along feature dimension.

The target agent feature $A_{target}$ is picked from $\mathcal{A}_{agents}$ and repeated $K$ times. Then it will be concatenated with agent query feature $Q^{j}_{agent}$ and map query feature $Q^{j}_{map}$ and finally goes through $\text{MLP}$ to generate the final feature for prediction.
\begin{equation}
    \begin{split}
            \mathcal{O}^{j} = \text{MLP} (
        [Q^{j}_{agent},Q^{j}_{map},A_{target}]
        ) \in \mathbb{R}^{K\times D}.
    \end{split}
\end{equation}

\textbf{Control Guided Multi-modal Motion Prediction.}
We follow\cite{shi_mtr_2022,varadarajan_multipath_2021} to utilize the Gaussian Mixture Model (GMM) to simulate the highly multi-modal distribution of agent future trajectories. Besides, to avoid infeasible trajectories, we additionally predict future control commands: acceleration $a$ and heading change rate $\dot{\theta}$ and then apply a kinematic model to transform control commands into trajectories. A regression head and classification head are utilized, respectively:
\begin{equation}
    \mathcal{P}^{j} = \text{MLP} (\mathcal{O}^{j}) \in \mathbb{R}^{K\times 1},
\end{equation}
\begin{equation}
    \Pi^{j} = \text{MLP} (\mathcal{O}^{j}) \in \mathbb{R}^{K\times (T\times 7)},
\end{equation}
where $\mathcal{P}^{j}$ is the predicted probability for each motion mode. For each future time step $t \in\{1,2,\cdots,T\}$, $\Pi^{j}_{t}=[	\mu_{x},\mu_{y},\sigma_{x},\sigma_{y},\rho,a,\dot{\theta}]^{j}_{K \times 7}$. We simply use the predicted centers $[\mu_{x},\mu_{y}]^{j}_{1:T}$ as the final prediction results $\Pi_{GMM}^{j} \in \mathbb{R}^{K\times (T\times 2)}$. For control commands $[a,\dot{\theta}]^{j}_{1:T}$, our kinematic model utilizes the Euler method to do discrete integration, as described in Algorithm \ref{algorithm-motion-update-vectorized}.

\begin{algorithm}[h]
    \caption{Kinematic Model}
    \label{algorithm-motion-update-vectorized}
    \begin{algorithmic}[1]
        \REQUIRE ~~\\ 
            Acceleration commands $\{ a_{1}, \dots, a_{T} \}$; \\
            Steering commands $\{ \dot{\theta}_{1}, \dots, \dot{\theta}_{T} \}$; \\
            Time step $\delta t$; \\
            Initial state $[x_{0}, y_{0}, \theta_{0}, v_{0}]$; \\
            Constraints: $a_{min}, a_{max}, \dot{\theta}_{min}, \dot{\theta}_{max}.$\\
        \ENSURE $\Pi_{control}$.
            \FORALL{$t \in \{ 1, \dots, T \}$}
                \STATE $v_{t} \gets v_{0} + \sum_{i=1}^{t} clamp(a_{i}, a_{min}, a_{max}) \cdot \delta t$
                \STATE $\theta_{t} \gets \theta_{0} + \sum_{i=1}^{t} clamp(\dot{\theta}_{i}, \dot{\theta}_{min}, \dot{\theta}_{max}) \cdot \delta t$
                \STATE $x_{t} \gets x_{0} + \sum_{i=1}^{t} (v_{i} \cdot \cos(\theta_{i}) \cdot \delta t)$
                \STATE $y_{t} \gets y_{0} + \sum_{i=1}^{t} (v_{i} \cdot \sin(\theta_{i}) \cdot \delta t)$
                \STATE $\Pi_{control}.insert([x_{t}, y_{t}])$
            \ENDFOR
            \RETURN $\Pi_{control}$
    \end{algorithmic}
\end{algorithm}

\subsection{Loss Function}
The model is trained in an end-to-end fashion using five different loss functions: (i) $\mathcal{L}_{dense}$, which is the auxiliary $L_{1}$ regression loss between $\Pi_{dense}$ and all agents' ground truth trajectories $\Pi_{allGT}$. (ii) $\mathcal{L}_{GMM}$, the GMM loss represented as negative log-likelihood loss for the predicted trajectories of the target agent $\Pi_{GMM}$. (iii) $\mathcal{L}_{cls}$, a classification loss in the form of cross-entropy loss on the predicted trajectory probability $\mathcal{P}$. (iv) $\mathcal{L}_{control}$, an $L_{1}$ regression loss between $\Pi_{control}$ and the target agents' ground truth $\Pi_{targetGT}$. (v) $\mathcal{L}_{guidance}$, an $L{1}$ regression loss between $\Pi_{control}$ and $\Pi_{GMM}$. $\mathcal{L}_{GMM},\mathcal{L}_{cls},\mathcal{L}_{control},\mathcal{L}_{guidance}$ are calculated as average sum of all decoder layers. Following \cite{varadarajan_multipath_2021,shi_mtr_2022}, we apply a hard-assignment strategy to optimize only one predicted trajectory whose corresponding intention point is closest to the ground truth endpoint.
The final loss is a weighted combination of these losses:
\begin{equation}
\begin{split}
    \mathcal{L}_{total} = & \lambda _{1}\mathcal{L}_{dense} + \lambda _{2}\mathcal{L}_{GMM} + \\
                          & \lambda _{3}\mathcal{L}_{cls} + \lambda _{4}\mathcal{L}_{control} + \lambda _{5}\mathcal{L}_{guidance},
\end{split}
\end{equation}
where $\lambda_{1,2,3,4,5}$ are hyper-parameters to balance the different loss terms.
Noted that $\mathcal{L}_{dense}, \mathcal{L}_{GMM},\mathcal{L}_{cls}$ are the same loss terms in baseline MTR model. We add $\mathcal{L}_{control}$ to learn how to generate control commands and add $\mathcal{L}_{guidance}$ to guide the $\Pi_{GMM}$ results.
% ,\mathcal{L}_{guidance}
%%%%%%%%%%%%%%%%%%%%%%%%%%%%%%%%%%%%%%%%%%%%%%%%%%%%%%%%%%%%%%%%%%%%%%%%%%%%%%%%

% TABLE I
\begin{table*}[tbp]
\centering
\caption{\href{https://waymo.com/open/challenges/2021/interaction-prediction/}{Joint Motion Prediction} Performance on Waymo Open Motion Dataset.}
\label{tab:joint}
\begin{tabular}{ccccccc} 
    \hline
    \rowcolor[rgb]{0.863,0.863,0.863} Method & Reference & minADE $\downarrow$ & minFDE $\downarrow$ & Miss Rate $\downarrow$ & Overlap Rate $\downarrow$ & Soft mAP $\uparrow$ \\ 
    \hline
    \\[-1em]
    Waymo LSTM baseline & Waymo & 1.9056 & 5.0278 & 0.7750 & 0.3407 & 0.0524 \\
    M2I & CVPR 2022 & 1.3506 & 2.8325 & 0.5538 & 0.2757 & 0.1239 \\
    GameFormer(M=64) & ICCV 2023 & 0.9721 & 2.2146 & 0.4933 & 0.2022 & 0.1982 \\
    MotionDiffuser & CVPR 2023 & \textbf{0.8642} & \textbf{1.9482} & \textbf{0.4300} & 0.2004 & 0.2047 \\
    MTR & NeurIPS 2022 & 0.9181 & 2.0633 & 0.4411 & \textbf{0.1717} & 0.2078 \\
    \rowcolor[rgb]{0.902,0.902,0.902} Ours & - & 0.9314 & 2.1079 & 0.4421 & \textbf{0.1717} & \textbf{0.2167} \\
    \hline
\end{tabular}
\end{table*}%

% TABLE II
\begin{table*}[tbp]
\centering
\caption{\href{https://waymo.com/open/challenges/2023/motion-prediction/}{Marginal Motion Prediction} Performance on Waymo Open Motion Dataset. Noticed that our writing is based on the results of our model for Waymo Challenge 2023, while the 2024 results are provided for reference.}
\label{tab:marginal}
\begin{tabular}{ccccccccc} 
    \hline
    \rowcolor[rgb]{0.863,0.863,0.863} Dataset~ & Method & Reference & minADE $\downarrow$ & minFDE $\downarrow$ & Miss Rate $\downarrow$& Overlap Rate $\downarrow$ & mAP $\uparrow$ & Soft mAP $\uparrow$ \\ 
    \hline
    \\[-1em]
    \multirow{6}{*}{Test~} 
     & DenseTNT & ICCV 2021 & 1.0387 & 1.5514 & 0.1573 & 0.1779 & 0.3281 & - \\
     & SceneTransformer & ICLR 2022 & 0.6117 & 1.2116 & 0.1564 & 0.1473 & 0.2788 & - \\
     & DIPP & TNNLS 2023 & 0.6951 & 1.4678 & 0.1854 &0.1516 &0.3383 & 0.3442 \\
     & ReCoAt & ITSC 2022 & 0.7703 & 1.6668 & 0.2437 & 0.1642& 0.2711 & - \\
     & HDGT & TPAMI 2023 & 0.5933 & 1.2055 & 0.1511 & 0.1557 & 0.3577 & 0.3709 \\
     & MotionCNN & Arxiv 2022 & 0.7400 & 1.4936 & 0.2091 & 0.1560& 0.2136 & - \\
     & MTR & NeurIPS 2022 & 0.6050 & 1.2207 & 0.1351 & 0.1277& 0.4129& 0.4216 \\
     & MTR++ & TRAMI 2024 & 0.5906 & 1.1939 & 0.1298 & 0.1281 &\textbf{0.4329} & 0.4410 \\
     & LLM-augmented MTR & Arxiv 2024 & 0.6181 & 1.2375 & 0.1402 & 0.1273 & 0.4025 & 0.4198 \\     
     & {\cellcolor[rgb]{0.902,0.902,0.902}}Ours &{\cellcolor[rgb]{0.902,0.902,0.902}}Waymo Challenge 2023 & {\cellcolor[rgb]{0.902,0.902,0.902}}\textbf{0.5904} & {\cellcolor[rgb]{0.902,0.902,0.902}}\textbf{1.1964} & {\cellcolor[rgb]{0.902,0.902,0.902}}\textbf{0.1295} & {\cellcolor[rgb]{0.902,0.902,0.902}} \textbf{0.1260} & {\cellcolor[rgb]{0.902,0.902,0.902}}0.4285 & {\cellcolor[rgb]{0.902,0.902,0.902}}\textbf{0.4436} \\ 
      & {\cellcolor[rgb]{0.902,0.902,0.902}}Ours & {\cellcolor[rgb]{0.902,0.902,0.902}}Waymo Challenge 2024 & {\cellcolor[rgb]{0.902,0.902,0.902}}0.5897 & {\cellcolor[rgb]{0.902,0.902,0.902}}1.1916 & {\cellcolor[rgb]{0.902,0.902,0.902}}0.1262 & {\cellcolor[rgb]{0.902,0.902,0.902}}0.1281 & {\cellcolor[rgb]{0.902,0.902,0.902}}0.4414 & {\cellcolor[rgb]{0.902,0.902,0.902}}0.4572 \\
    \hline
    \\[-1em]
    \multirow{2}{*}{Validation~} & MTR & NeurIPS 2022 & 0.6046 & 1.2251 & 0.1366 &- &0.4164 & - \\
    % & MTR++ & TRAMI 2024 & 0.5912 & 1.1986 & 0.1296 & -& \textbf{0.4351}&-  \\
     & {\cellcolor[rgb]{0.902,0.902,0.902}}Ours & {\cellcolor[rgb]{0.902,0.902,0.902}}Waymo Challenge 2023 & {\cellcolor[rgb]{0.902,0.902,0.902}}\textbf{0.5900} & {\cellcolor[rgb]{0.902,0.902,0.902}}\textbf{1.2010} & {\cellcolor[rgb]{0.902,0.902,0.902}}\textbf{0.1323} & {\cellcolor[rgb]{0.902,0.902,0.902}}\textbf{0.1267} & {\cellcolor[rgb]{0.902,0.902,0.902}}\textbf{0.4220} & {\cellcolor[rgb]{0.902,0.902,0.902}}\textbf{0.4373} \\
    & {\cellcolor[rgb]{0.902,0.902,0.902}}Ours & {\cellcolor[rgb]{0.902,0.902,0.902}}Waymo Challenge 2024 & {\cellcolor[rgb]{0.902,0.902,0.902}}0.5926 & {\cellcolor[rgb]{0.902,0.902,0.902}}1.2059 & {\cellcolor[rgb]{0.902,0.902,0.902}}0.1318 & {\cellcolor[rgb]{0.902,0.902,0.902}}0.1267 & {\cellcolor[rgb]{0.902,0.902,0.902}}0.4413 & {\cellcolor[rgb]{0.902,0.902,0.902}}0.4581 \\
        \hline
\end{tabular}
\end{table*}

\section{EXPERIMENTS}

\begin{figure*}[t]
    \centering
    \includegraphics[width=15.7cm]{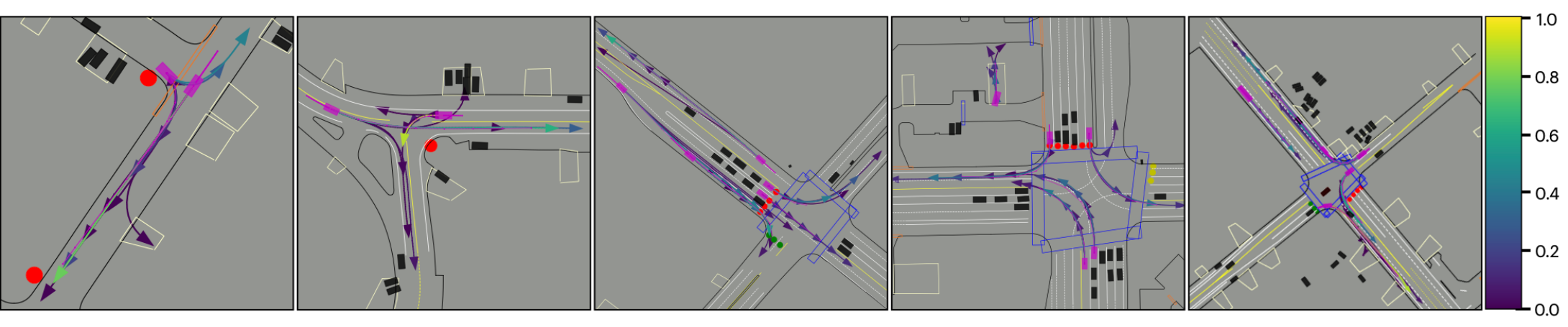} 
    \caption{
        Qualitative evaluation of the top-$6$ marginal prediction results in different scenarios. The colour of each trajectory reflects its predicted probability according to the colour bar on the right.
        }
    \label{fig:marginal}
    \vspace{-1em}
\end{figure*}

\begin{figure*}[t]
    \centering
    \includegraphics[width=15.7cm]{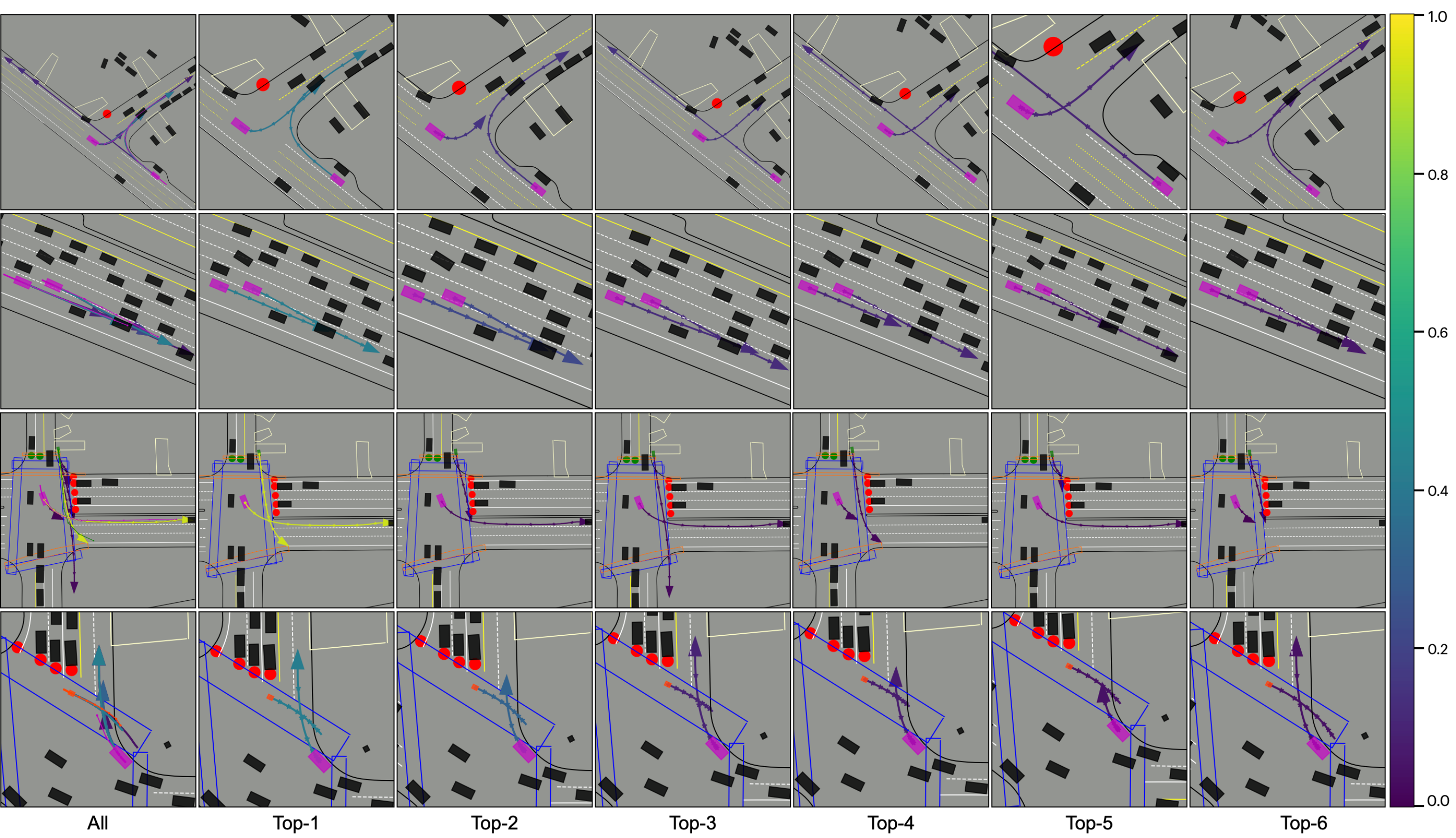} 
    \caption{
        % \textcolor{blue}{
        Qualitative evaluation of the top-$6$ joint prediction results in different scenarios. The colour of each trajectory reflects its predicted probability according to the colour bar on the right. The black, purple, green and orange colour rectangles represent other vehicles, target vehicles, cyclists and pedestrians, respectively. The ground truth trajectories are drawn with the same color together on the first column.  
        % }
    }
    \label{fig:joint}
    \vspace{-1em}
\end{figure*}
\subsection{Experimental Setup}
\subsubsection{Dataset and metrics} 
% \textcolor{red}{To evaluate the efficacy of the proposed approach, we employed the {\bf Waymo Open Motion Dataset}. This dataset stands out due to its extensive compilation of annotated objects, showcasing interactive behaviors across a myriad of road configurations. Notably, our experimentation is based on its most recent iteration, version 1.2, enriched with road connection data. For a structured analysis, the dataset delineates distinct training, validation, and test segments. Furthermore, it boasts an online leaderboard that facilitates an objective assessment of algorithmic performance against the test partition. Within this dataset, the focal agents are systematically classified into three primary categories: vehicles, pedestrians, and cyclists.}

% TABLE III
\begin{table*}[tbp]
\caption{Ablation Study on Proposed Modules.}
\label{tab:ablation}%
\centering
\begin{tabular}{cccccccccc} 
    \hline
    \rowcolor[rgb]{0.863,0.863,0.863} Method & \begin{tabular}[c]{@{}>{\cellcolor[rgb]{0.863,0.863,0.863}}c@{}}Scene-compliant \\Intention Points\end{tabular} & \begin{tabular}[c]{@{}>{\cellcolor[rgb]{0.863,0.863,0.863}}c@{}}Control \\Guidance\end{tabular} & \begin{tabular}[c]{@{}>{\cellcolor[rgb]{0.863,0.863,0.863}}c@{}}MSG \\Encoder\end{tabular} & MCG & Soft mAP $\uparrow$ & minADE $\downarrow$ & minFDE $\downarrow$ & Miss Rate $\downarrow$ & \begin{tabular}[c]{@{}>{\cellcolor[rgb]{0.863,0.863,0.863}}c@{}}Cross-boundary \\Rate ($K=64$)$\downarrow$\end{tabular} \\ 
    \hline
    \\[-1em]
    MTR & \XSolidBrush & \XSolidBrush & \XSolidBrush & \XSolidBrush & 0.4218 & 0.6114 & 1.3480 & 0.1348 & 0.6613 \\ 
    \hline
    \\[-1em]
    \multirow{4}{*}{Ours} & \checkmark & \XSolidBrush & \XSolidBrush & \XSolidBrush & 0.4225 & 0.6059 & 1.2257 & 0.1350 & - \\
     & \checkmark & \checkmark & \XSolidBrush & \XSolidBrush & 0.4307 & 0.6039 & 1.2209 & 0.1338 & - \\
     & \checkmark & \checkmark & \checkmark & \XSolidBrush & 0.4312 & 0.5980 & 1.2127 & 0.1330 & - \\
     & {\cellcolor[rgb]{0.902,0.902,0.902}}\checkmark & {\cellcolor[rgb]{0.902,0.902,0.902}}\checkmark & {\cellcolor[rgb]{0.902,0.902,0.902}}\checkmark & {\cellcolor[rgb]{0.902,0.902,0.902}}\checkmark & {\cellcolor[rgb]{0.902,0.902,0.902}}\textbf{0.4324} & {\cellcolor[rgb]{0.902,0.902,0.902}}\textbf{0.5923} & {\cellcolor[rgb]{0.902,0.902,0.902}}\textbf{1.2018} & {\cellcolor[rgb]{0.902,0.902,0.902}}\textbf{0.1326} & {\cellcolor[rgb]{0.902,0.902,0.902}}\textbf{0.3868} \\
    \hline
\end{tabular}
\end{table*}

Our model is trained and evaluated based on the Waymo Open Motion Dataset (WOMD). 
WOMD introduces two challenges: the marginal motion prediction challenge, which independently assesses the predicted motion of target agents ($2$ to $8$ agents in each scenario), and the joint motion prediction challenge, which necessitates predicting joint future positions of $2$ interacting agents. 
Both challenges share the same dataset split, comprising $487$k scenes for training, $44$k for validation, and $44$k for testing. For training and validation, each scene consists of $1$ second of historical agents' tracks and $8$ second future tracks, requiring the marginal or joint prediction for $8$ second future. We assess model performance using metrics including minimum average displacement error (minADE), minimum final displacement error (minFDE), miss rate, overlap rate, and soft mean average precision (Soft mAP), with Soft mAP serving as pivotal indicators for evaluating the model's performance. Especially we propose a cross-boundary rate, which is defined as cross-boundary predictions divided by all predictions. The boundaries we consider here refer to solid double yellow lines, solid double white lines, road edge boundaries and road edge medians. 

\begin{figure}[h]
    \centering
    \includegraphics[width=8cm]{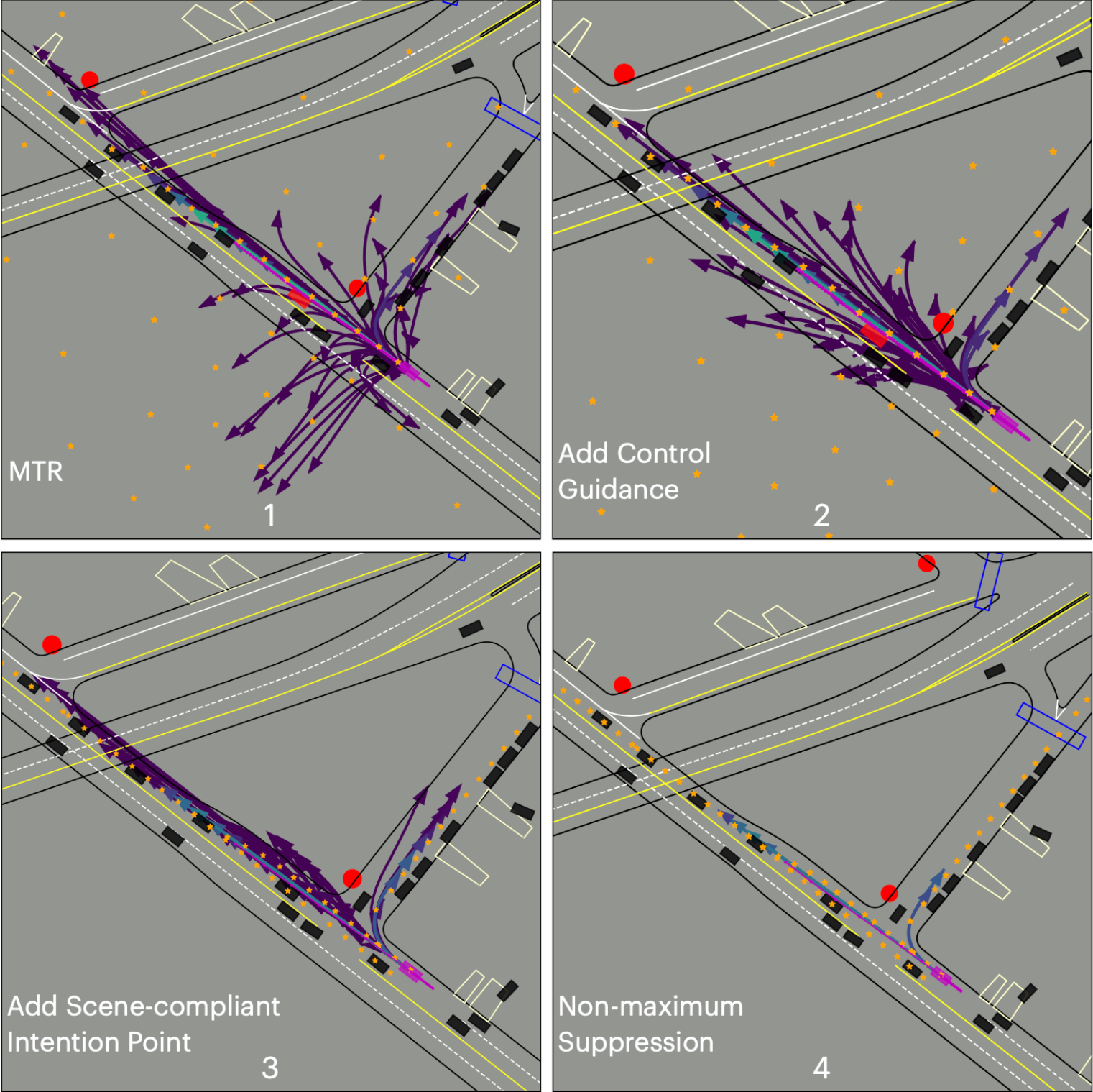} 
    \caption{
        Qualitative results of our proposed modules. After adding control guidance and applying scene-compliant intention points, the distribution of the prediction results is confined within driverable area. Finall we use NMS to select the best top-6 trajectories.
        }
    \label{fig:overflow}
    \vspace{-1em}
\end{figure}

\subsubsection{Implementation details} The training of our model is conducted end-to-end using the AdamW optimizer. Key parameters include a learning rate initialized as 0.0001, weight decay at 0.01, and a batch size of 48. Utilizing 8 NVIDIA-3090 GPUs, the model undergoes a training period spanning 35 epochs. Commencing from epoch 20, the learning rate experiences a decay by a factor of 0.5 every two epochs and remains constant after 30 epochs. We stack two layers of MCG and four layers of local-attention transformers in the encoder, and six cross-attention layers in the decoder. For each target agent, we predict 64 potential trajectories using 64 prior intention points. Then Non-maximum suppression (NMS) technique is utilized to generate six trajectories as final prediction results. For loss coefficient， we set $\lambda_{1,2,3,4}=1$ and $\lambda_{5}=0.1$. Other training details are identical to the original MTR unless stated otherwise.

\subsection{Joint Motion Prediction Performance}
Following previous studies \cite{sun2022m2i, shi_mtr_2022}, we evaluate our joint motion prediction methodology by merging the marginal predictions of two interacting agents into a joint prediction. TABLE \ref{tab:joint} outlines the joint motion prediction performance on WOMD, presenting a comparison of various methods across key reference metrics. Our method surpasses other approaches in both Overlap Rate and Soft mAP metrics, with a particularly notable superiority in Soft mAP, a pivotal metric for assessing performance. Further qualitative results are depicted in Fig. \ref{fig:joint}, illustrating the effectiveness of our method in complex interacting scenarios.

\subsection{Marginal Motion Prediction Performance}
The performance of various marginal motion prediction methodologies on WOMD is detailed in TABLE \ref{tab:marginal}. When evaluated on the test dataset, our method surpasses MTR on all metrics and demonstrates superior performance across most metrics compared with other methods. In particular, we achieve the highest Soft mAP, 0.4436. Our Method consistently outperforms MTR and its successor, MTR++, across all metrics in the validation dataset. Additional qualitative results are provided in Fig. \ref{fig:marginal}. 

% \textcolor{red}{More improvements are observed in minADE and minFDE, with our method achieving 0.5923 and 1.2018, respectively, compared to MTR's 0.6046 and 1.2251.
% The Soft mAP for our method is 0.4324, surpassing MTR's 0.4216, indicating consistent performance superiority in accurate motion prediction. Additional qualitative results are provided in Fig. \ref{fig:marginal}.} 
% In summary, the comparative analysis demonstrates that Our Method, building upon and enhancing the MTR framework, exhibits significant improvements across all metrics, particularly in critical metrics like Miss Rate and mAP. This leads to an overall enhanced performance in both test and validation scenarios, indicating that these refinements have effectively optimized motion prediction capabilities. 

\subsection{Ablation Study}
We study the effects of individual modules and the control guidance weight on the performance of our pipeline. TABLE \ref{tab:ablation} is generated using the entire training and testing dataset, while TABLE \ref{tab:ablation-cg} is based on a randomly sampled 20\% subset of the training dataset and full validation dataset. Noted that all ablation studies are trained 30 epochs for efficiency.
\subsubsection{Effects of the proposed modules}
TABLE \ref{tab:ablation} illustrates the ablation analysis of scene-compliant intention points, control guidance(CG), MSG, and MCG and Figure \ref{fig:overflow} shows the qualitative results of our proposed modeuls.

The integration of scene-compliant intention points and CG significantly enhances the model's performance. Nearly every metric indicates an improvement compared to the previous variant upon the introduction of these modules.

MSG Encoder:
Adding the MSG Encoder to the model leads to further improvements. There is a slight increase in Soft mAP to 0.4312. More significantly, this addition results in decreased minADE (0.6039 to 0.5980), minFDE (1.2209 to 1.2127), and Miss Rate (0.1338 to 0.1330). These outcomes indicate that the MSG Encoder synergizes well with the previous modules, enhancing the model's overall performance.

MCG:
The integration of MCG marks the highest performance improvement in our ablation scenarios. The Soft mAP value reaches its peak at 0.4324. Concurrently, there is a reduction in minADE to 0.5923, minFDE to 1.2018, and Miss Rate to 0.1326. 
Furthermore, to demonstrate the necessity of the MCG Layer that accounts for global attention, we conducted an ablation study as shown in TABLE \ref{tab:mcg_layer}. The result revealed that the model achieves its best performance in terms of minADE, minFDE, and Miss Rate when the MCG Layer is set to 2.

Additionally, a significant improvement is observed in the Cross-boundary Rate, which drops to 0.3868, showing that our model can learn map constraint information and output on-road yet feasible trajectories.
\begin{table}[htb]
\caption{Ablation Study on MCG Layer}
\centering
\begin{tabular}{cccc}
\hline 
\rowcolor[rgb]{0.863,0.863,0.863}
MCG Layer & minADE~$\downarrow$ & minFDE~$\downarrow$ & MissRate~$\downarrow$ \\
\hline
0 & 0.6633 & 1.3648 & 0.1631 \\
2 & \textbf{0.6561} & \textbf{1.3441} & \textbf{0.1602} \\
4 & 0.6592 & 1.3598 & 0.1628 \\
\hline
\end{tabular}

\label{tab:mcg_layer}
\end{table}

\subsubsection{Effects of control guidance weight}
% TABLE IV
\begin{table}[htb]
\centering
\caption{Ablation Study on Control Guidance (CG) Weight $\lambda_{5}$.}
\label{tab:ablation-cg}
\begin{tabular}{cccccc} 
    \hline
    \rowcolor[rgb]{0.863,0.863,0.863} \begin{tabular}[c]{@{}>{\cellcolor[rgb]{0.863,0.863,0.863}}c@{}}CG \\Weight\\\end{tabular} & minADE~$\downarrow$ & minFDE~$\downarrow$ & \begin{tabular}[c]{@{}>{\cellcolor[rgb]{0.863,0.863,0.863}}c@{}}Miss \\Rate~$\downarrow$\end{tabular} & \begin{tabular}[c]{@{}>{\cellcolor[rgb]{0.863,0.863,0.863}}c@{}}Overlap \\Rate~$\downarrow$\end{tabular} & \begin{tabular}[c]{@{}>{\cellcolor[rgb]{0.863,0.863,0.863}}c@{}}Soft \\mAP~$\uparrow$\end{tabular} \\ 
    \hline
    \\[-1em]
    0.0  & 0.6737 & 1.3725 & 0.1654 & \textbf{0.1341} & 0.3616 \\
    \rowcolor[rgb]{0.902,0.902,0.902}0.1 & \textbf{0.6602} & \textbf{1.3497} & \textbf{0.1615} & 0.1346 & \textbf{0.3679} \\ 0.2 & 0.6787 & 1.3725 & 0.1627 & 0.1361 & 0.3645 \\
    0.3 & 0.6833 & 1.3937 & 0.1700 & 0.1384 & 0.3436 \\
    0.4 & 0.7013 & 1.3916 & 0.1705 & 0.1366 & 0.3473 \\

    \hline
\end{tabular}
\end{table}
In our ablation study, evaluating varied CG weights on model performance (TABLE \ref{tab:ablation-cg}), a CG weight of 0.1 notably yields the highest Soft mAP (0.3679) and the lowest Miss Rate (0.1615), demonstrating the effectiveness of the proposed control guidance. 
\begin{figure}[ht]
    \centering
    \includegraphics[width=8.5cm]{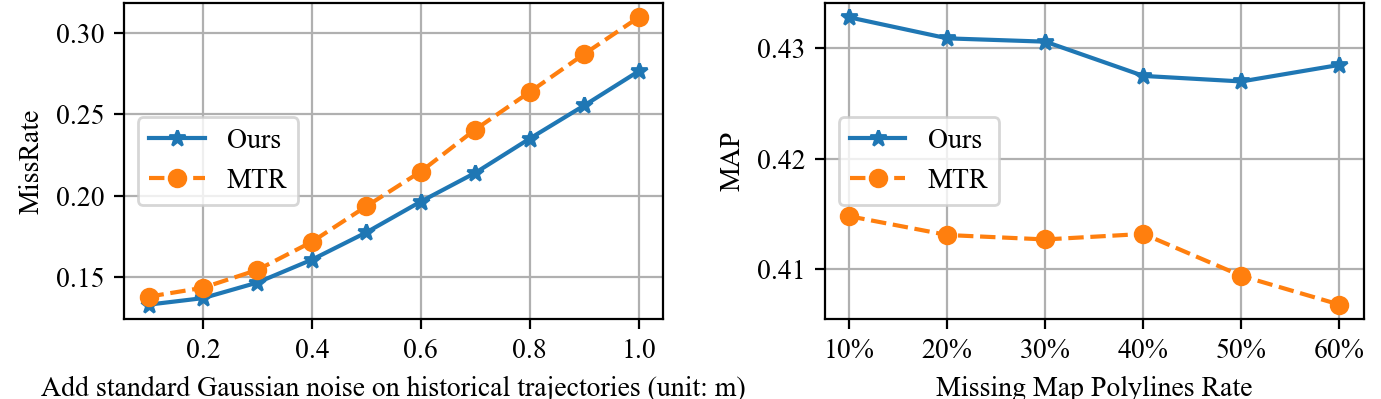} 
    \caption{Robustness between our model and MTR baseline.}
  
    \label{fig:Robustness}
    \vspace{-1em}
\end{figure} 
\subsection{Model Robustness Analysis}
Trajectory prediction performance is usually influenced by noisy past observations. Additionally, in scenarios such as those involving road construction or non-structural roads, some map polylines may be missing. To simulate these cases and test our model's robustness, we add standard Gaussian noise on past trajectories and randomly mask some map polylines. Fig. \ref{fig:Robustness} shows that our model exhibits less increase in MissRate and maintains strong
 MAP performance as the missing map polylines rate increases, showcasing its robustness compared with the baseline MTR model.

\section{CONCLUSIONS}
In this paper, we propose {\method} framework,
which applies traffic-rule-compliant intention points and generates additional control commands to guide the direct predictions. An innovative scene encoder architecture is presented for better modelling of the intricate relationships between agents and map polylines. Our method can significantly decrease off-road and infeasible predictions while maintaining even more accurate prediction metrics. In the future, we plan to design a lightweight structure to directly learn the possible intention points, and deploy our algorithm on the real vehicle.

\addtolength{\textheight}{-4cm}   % This command serves to balance the column lengths

%%%%%%%%%%%%%%%%%%%%%%%%%%%%%%%%%%%%%%%%%%%%%%%%%%%%%%%%%%%%%%%%%%%%%%%%%%%%%%%%

%%%%%%%%%%%%%%%%%%%%%%%%%%%%%%%%%%%%%%%%%%%%%%%%%%%%%%%%%%%%%%%%%%%%%%%%%%%%%%%%



% \section*{ACKNOWLEDGMENT}

%%%%%%%%%%%%%%%%%%%%%%%%%%%%%%%%%%%%%%%%%%%%%%%%%%%%%%%%%%%%%%%%%%%%%%%%%%%%%%%%

\bibliographystyle{IEEEtran}
\bibliography{root}

\end{document}